\documentclass[11pt]{article}

\usepackage{acl}
\usepackage{times}
\usepackage{latexsym}
\usepackage[T1]{fontenc}
\usepackage[utf8]{inputenc}
\usepackage{microtype}
\usepackage{inconsolata}
\usepackage{graphicx}
\usepackage{amsmath}
\usepackage{booktabs}
\usepackage{tabularx}
\usepackage{multirow}
\usepackage{colortbl}
\usepackage[most]{tcolorbox}
\tcbuselibrary{listings}

\title{Heading Hierarchy Reconstruction for Complex and Long Documents with Level-Wise Multimodal Memory}
\author{Qian Chen, Xianyin Zhang, Lifan Guo, Chi Zhang \\
  Qwen DianJin Team, Alibaba Cloud Computing \\
  \texttt{\{xiaoyu.cq,zhangxianyin.zxy,lifan.lg,edward.zhang\}@alibaba-inc.com}}

\begin{document}
\maketitle
\begin{abstract}
Vision-language models (VLMs) have been widely adopted for document parsing and optical character recognition (OCR), as they can accurately extract page-level elements. However, downstream applications such as retrieval-augmented generation (RAG) also require coherent document-level structure. Although these models indicate heading levels using different numbers of preceding ``\#'' symbols, page-by-page prediction often lacks document-level consistency. In practice, input-resolution and context-length constraints make it infeasible for a VLM to jointly process all page images in documents containing tens or hundreds of pages. To address this limitation, we propose a VLM-based framework that incrementally reconstructs heading hierarchies with a \emph{level-wise multimodal memory}. The memory retains a compact subset of all previously processed pages, selected by anchors. Our framework supports flexible anchor selection, and in this work we select the most recent node at each level, allowing the VLM to observe the visual, textual, and layout patterns of headings at every level. For comprehensive evaluation, we also introduce the Complex and Long Document Heading Hierarchy Reconstruction Benchmark (CLDoc-HHRBench), comprising 250 documents and 11,212 pages with expert-verified annotations. Experiments on CLDoc-HHRBench and three existing benchmarks demonstrate the effectiveness of our method.
\end{abstract}


\section{Introduction}
\label{sec:intro}
From scientific papers and technical manuals to laws and business reports, a massive amount of knowledge is stored in unstructured documents. To effectively manage, retrieve, and understand these documents, document parsing has become a critical research topic. Among the structural components of a document, its heading hierarchy provides a navigational backbone: it organizes section headings into a coherent tree and associates them with their locations in the document. Recovering this hierarchy is important for downstream applications such as retrieval-augmented generation (RAG), where document structure supports structure-aware chunking, navigation, and full-text understanding. In this work, we focus specifically on \emph{heading hierarchy reconstruction}, the task of assigning globally consistent hierarchy levels to document headings and recovering their tree-structured organization.

\begin{figure*}[tb]
  \centering
  \includegraphics[width=1.0\textwidth]{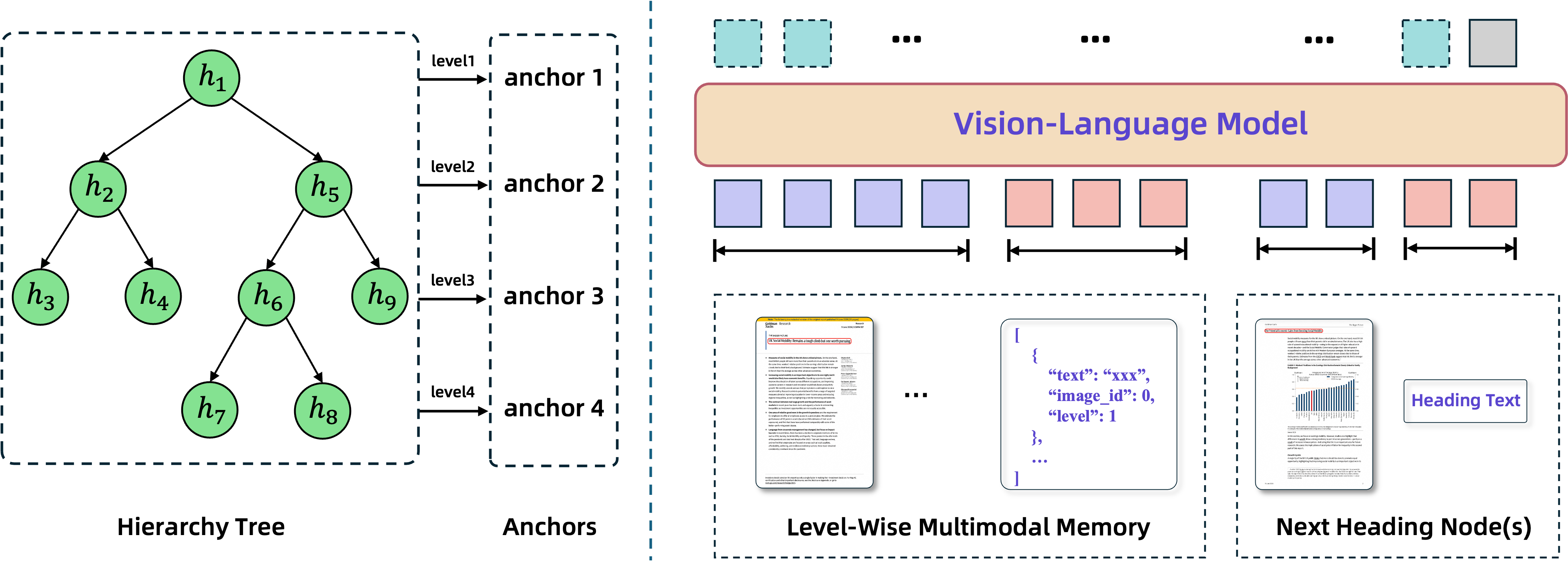}
  \caption{An overview of our proposed method. \textbf{Left}: a hierarchy tree constructed from the preceding predictions, where each circle denotes a heading node. \textbf{Right}: the memory and the next target heading are fed into the VLM, which is prompted to predict the target heading's hierarchy level. Anchors guarantee level-wise coverage, while other headings on the selected pages provide additional references for prediction.}
  \label{fig:example}
\end{figure*}

Heading hierarchy reconstruction remains challenging because document layouts, heading styles, and hierarchy depths vary widely. Early approaches relied on heuristic rules and regular expressions, while subsequent machine- and deep-learning methods combined CNNs, BERT \cite{devlin2019bert}, and optical character recognition (OCR) engines to model visual, textual, and layout features. However, the reliance of these methods on multiple specialized components and limited-scale training data can hinder cross-modal alignment and generalization. Recently, methods based on vision-language models (VLMs) have reconstructed heading hierarchies from page-level parsing results. MinerU-Popo \cite{xu2026mineru} uses dynamic chunking, vertically concatenates pages within each chunk, and synchronizes levels through overlaps. The resulting long images may dilute per-page visual detail, while offset-based synchronization cannot directly correct non-uniform within-chunk errors. LingDT-VL-OCR \cite{qian2026lingdtvlocr} aggregates heading crops into Pseudo-ToC images and prompts a VLM to recover heading levels. As the number of headings increases, however, both the Pseudo-ToC input and the generated hierarchy become longer, making predictions more susceptible to omissions, duplications, and inconsistent levels in ultra-long documents.

These limitations reveal a fundamental trade-off when applying VLMs to long documents. Providing more page images preserves fine-grained visual evidence, but tens or hundreds of images quickly exceed the input-resolution and context-length limits of a VLM. Combining many pages into a single image reduces the image count, but may compress page-level visual cues. To balance visual fidelity and historical coverage, we ask whether heading hierarchies can be reconstructed without repeatedly providing the complete page history, while still preserving sufficient historical context. To answer this question, we propose a VLM-based framework that incrementally reconstructs heading hierarchies with a \emph{level-wise multimodal memory}. The memory uses heading nodes as anchors to select a compact set of previously processed pages and supports flexible anchor-selection strategies. In this work, we retain the most recent node at each observed hierarchy level. The selected pages therefore preserve representative visual styles, heading texts, layout patterns, and predicted structural relationships across all levels encountered so far. At each prediction step, these historical references are provided together with the next target heading or target pages, enabling the VLM to infer new levels while maintaining document-level consistency. Because the historical memory remains fixed within each inference step, the framework supports prediction at different granularities, ranging from a single heading to headings on multiple consecutive pages.

In addition, existing benchmarks do not adequately capture both the length and structural complexity of real-world documents. Comp-HRDoc \cite{wang2024detect} consists solely of scientific papers with similar layouts and hierarchical structures. Although DocHieNet \cite{xing2024dochienet} covers multiple document types, its document length and average hierarchy complexity remain limited. To support the evaluation of heading hierarchy reconstruction on complex and long documents, we introduce the Complex and Long Document Heading Hierarchy Reconstruction Benchmark (CLDoc-HHRBench). It comprises 250 documents and 11,212 pages across 10 document categories, covering diverse layouts and larger heading hierarchies than existing benchmarks. Its longest document spans 387 pages, while some documents contain hundreds or even over a thousand headings.

Our main contributions can be summarized as follows:
\begin{itemize}
    \item We propose a VLM-based framework with a \emph{level-wise multimodal memory} for heading hierarchy reconstruction.
    \item We introduce CLDoc-HHRBench, a benchmark covering complex, long, and diverse documents.
    \item Extensive experiments validate our method and provide a comprehensive evaluation of heading hierarchy reconstruction on CLDoc-HHRBench.
\end{itemize}

\section{Related Work}

\subsection{Vision-Language Models}

\subsubsection{General VLMs}
General VLMs have emerged as an alternative paradigm for document parsing and understanding. Closed-source models such as the GPT series \cite{achiam2023gpt} and the Gemini family \cite{team2023gemini} demonstrate strong OCR capabilities, often outperforming traditional pipeline-based systems in recognizing text, formulas, and tables. Open-source models such as the Qwen collection \cite{wang2024qwen2, bai2025qwen3} and the InternVL family \cite{zhu2025internvl3, wang2025internvl35} also show considerable promise for OCR applications.

\subsubsection{OCR VLMs}
OCR-oriented VLMs simplify document parsing by directly processing entire document pages or substantial regions through a unified framework, generating structured outputs without the need for multiple specialized models. End-to-end methods like DeepSeek-OCR \cite{wei2025deepseek} and FireRed-OCR \cite{wu2026firered} unify recognition within a single framework, while two-stage methods like PaddleOCR-VL \cite{cui2025paddleocr} and MinerU 2.5 \cite{niu2026mineru2} first perform page-level layout analysis and then conduct efficient parallel parsing of identified regions. Although these models excel at layout analysis, reading order detection, and multiple element recognition, they do not explicitly address document-level heading hierarchy reconstruction.

\subsection{Heading Hierarchy Reconstruction Methods}

\subsubsection{Rule-based Methods}
Rule-based methods \cite{belaid2001recognition,dresevic2008book,marinai2010table,wu2013table} have been widely adopted, relying on predefined patterns, hand-crafted rules, or algorithmic heuristics. These methods typically follow a common pipeline: first identifying the pages that contain the table of contents, then extracting individual entries, and finally linking each entry to its corresponding section in the document.
\subsubsection{ML-based/DL-based Methods} 

In contrast to rule-based methods, machine learning (ML) and deep learning (DL) methods \cite{dejean2009tables, klampfl2014unsupervised, bentabet2020automatic, wang2024detect} introduce greater flexibility by leveraging textual, visual, and layout features. D-O-C \cite{wang2024detect} encodes section headings with a Transformer, predicts parent--child and sibling scores, and serially inserts each heading as the rightmost child of a selected node on the current rightmost subtree.

\subsubsection{VLM-based Methods}
Benefiting from strong cross-modal capabilities, VLMs can jointly interpret the visual appearance, textual semantics, and spatial layout of document headings. MinerU-Popo \cite{xu2026mineru} processes long documents through dynamic chunking and synchronization, while LingDT-VL-OCR \cite{qian2026lingdtvlocr} aggregates heading crops into Pseudo-ToC images for document-level hierarchy reconstruction.

\subsection{Heading Hierarchy Reconstruction Benchmarks}
Early benchmarks for heading hierarchy reconstruction were predominantly constructed from arXiv papers. arXivdocs \cite{rausch2021docparser} is an early benchmark for hierarchical structure construction and contains 362 annotated pages. HRDoc \cite{ma2023hrdoc} aims to recover the hierarchical structure of PDF documents and contains 2,500 multi-page documents. Comp-HRDoc \cite{wang2024detect} is built upon the HRDoc-Hard dataset, which comprises 1,000 documents for training and 500 documents for testing. Nevertheless, these datasets are limited to scientific articles with similar structures. More recent benchmarks incorporate more complex layouts and a broader range of document types. E-Periodica \cite{rausch2023dsg} mitigates this homogeneity and comprises 542 single pages from magazines. DocHieNet \cite{xing2024dochienet} contains 1,673 documents from domains including legal, financial, educational, technical, and scientific materials. MPDocBench-Parse \cite{zhou2026mpdocbench} contains 433 manually annotated documents with 3,246 pages across 15 document types (of which 420 documents with 3,135 pages are publicly released). However, their document length and hierarchy depth remain limited.

\section{Method}
We incrementally reconstruct a document's heading hierarchy using a VLM equipped with a level-wise multimodal memory, as illustrated in Figure~\ref{fig:example}. This section first defines the task and then describes the memory construction and multimodal level prediction.

\subsection{Problem Definition}
Given a document $D$, we assume that its section headings have been detected, placed in reading order, and transcribed by upstream processing, yielding a heading sequence
\begin{equation}
    H = (h_1, h_2, \ldots, h_n).
\end{equation}
Each heading is represented as
\begin{equation}
    h_i = (I_{p(h_i)}, t_i, b_i),
\end{equation}
where $p(h_i)$ denotes the index of the page containing $h_i$, $I_{p(h_i)}$ is the corresponding page image, $t_i$ is the heading text, and $b_i=(x_i^{\min},y_i^{\min},x_i^{\max},y_i^{\max})$ is its bounding box on that page. Before a page image is provided to the VLM, the corresponding heading bounding box is drawn directly on the image. The marked page preserves the heading's authentic visual and layout context while explicitly identifying the referenced heading. The numerical coordinates are not serialized in the textual prompt.

Heading hierarchy reconstruction aims to predict a hierarchy level
\begin{equation}
    l_i\in\{1, 2, \ldots, L\},
\end{equation}
for every heading, forming an ordered heading--level sequence
\begin{equation}
    S=((h_1,l_1),(h_2,l_2),\ldots,(h_n,l_n)),
\end{equation}
where $L$ denotes the maximum hierarchy depth. Given $S$, the hierarchy tree can be constructed deterministically: processing the headings in reading order, each heading with level $l_i>1$ is attached as a child of the most recent preceding heading with level $l_i-1$, and each heading with $l_i=1$ is attached to a virtual root node.

\subsection{Level-Wise Multimodal Memory}
To preserve document-level context within the VLM input limits, we construct a level-wise multimodal memory.

Specifically, after processing the first $i$ headings, let $d_i$ denote the maximum hierarchy level in the hierarchy tree constructed so far:
\begin{equation}
    d_i=\max_{1\leq j\leq i}\hat{l}_j,
\end{equation}
where $\hat{l}_j$ denotes the predicted hierarchy level of $h_j$. We introduce one anchor for each hierarchy level present in the constructed tree and denote the anchor set by
\begin{equation}
    A_i=\left\{a_i^{(1)},a_i^{(2)},\ldots,a_i^{(d_i)}\right\},
\end{equation}
where $a_i^{(k)}$ is the representative heading node assigned level $k$. The historical pages containing these anchors constitute the selected page set
\begin{equation}
    \mathcal{P}_i=\left\{p(h_j)\mid h_j\in A_i\right\}.
\end{equation}
Using $\mathcal{P}_i$, the multimodal memory is defined as
\begin{equation}
    M_i=\left\{\left(I_q,\left\{(t_j,b_j,\hat{l}_j)\right\}_{\substack{1\leq j\leq i\\p(h_j)=q}}\right)\mid q\in\mathcal{P}_i\right\}.
\end{equation}
When multiple anchors lie on the same page, that page is selected only once. Anchors are used solely for memory-page selection, and all processed headings on the selected pages remain available as structural references. Accordingly, besides the anchors, the VLM input includes the text, bounding boxes, and predicted levels of all other processed headings on those pages. Since there is one anchor for every hierarchy level in the constructed tree, $|\mathcal{P}_i|\leq |A_i|=d_i$. Thus, at each inference step, the VLM receives only the historical pages selected by the anchors rather than the complete set of previously processed pages. The bound characterizes the number of selected historical pages in terms of the depth $d_i$ of the currently constructed tree.

The memory is initialized with the first page that contains at least one heading. We assign level 1 to the first heading and treat the remaining headings in reading order as subsequent prediction targets. Our method directly predicts the level of each heading and remains applicable even if the document contains additional level-1 headings. In contrast, some prior relation-prediction methods \cite{wang2024detect} treat the first heading as the root and cannot assign a later heading to the same level.

The framework is not tied to a specific anchor-selection strategy. Different strategies can select representative nodes for the observed levels while retaining the same memory construction and prediction procedure. In this work, we select the most recent node predicted at each level as the anchor, yielding a deterministic strategy that maintains one reference for every observed level. For example, when predicting the level of ``Section 2.2,'' ``Chapter 2'' is more informative than ``Chapter 1,'' while ``Section 2.1'' is more informative than ``Section 1.1'' or ``Section 1.2.''

\begin{figure*}[!t]
  \begin{minipage}[t]{0.49\textwidth}
    \centering
    \includegraphics[width=0.92\linewidth]{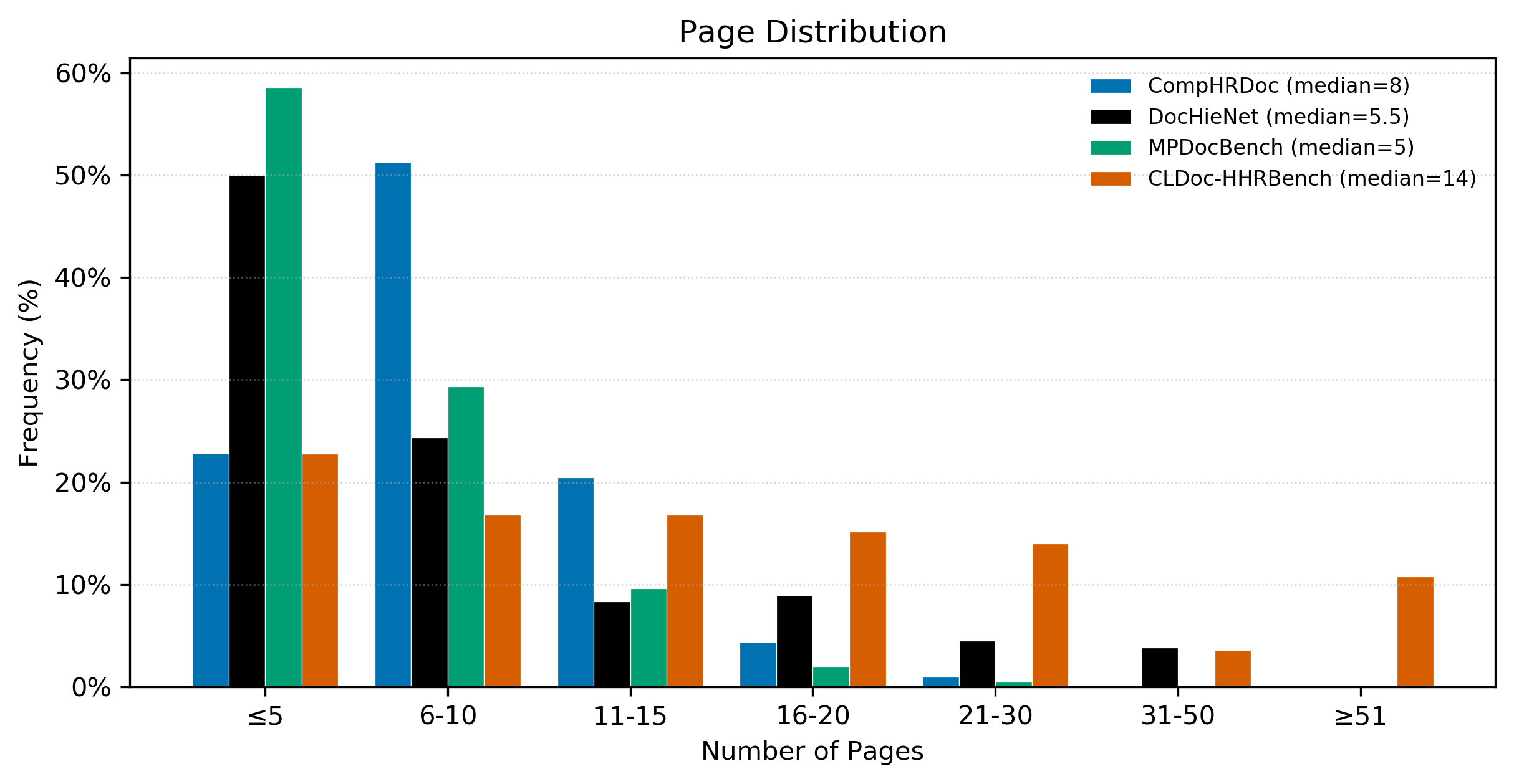}
    \captionof{figure}{Distribution of document lengths (pages containing headings) across benchmarks. CLDoc-HHRBench has the highest median and a larger proportion of long documents.}
    \label{fig:page-distribution}
  \end{minipage}
  \hfill
  \begin{minipage}[t]{0.49\textwidth}
    \centering
    \includegraphics[width=0.92\linewidth]{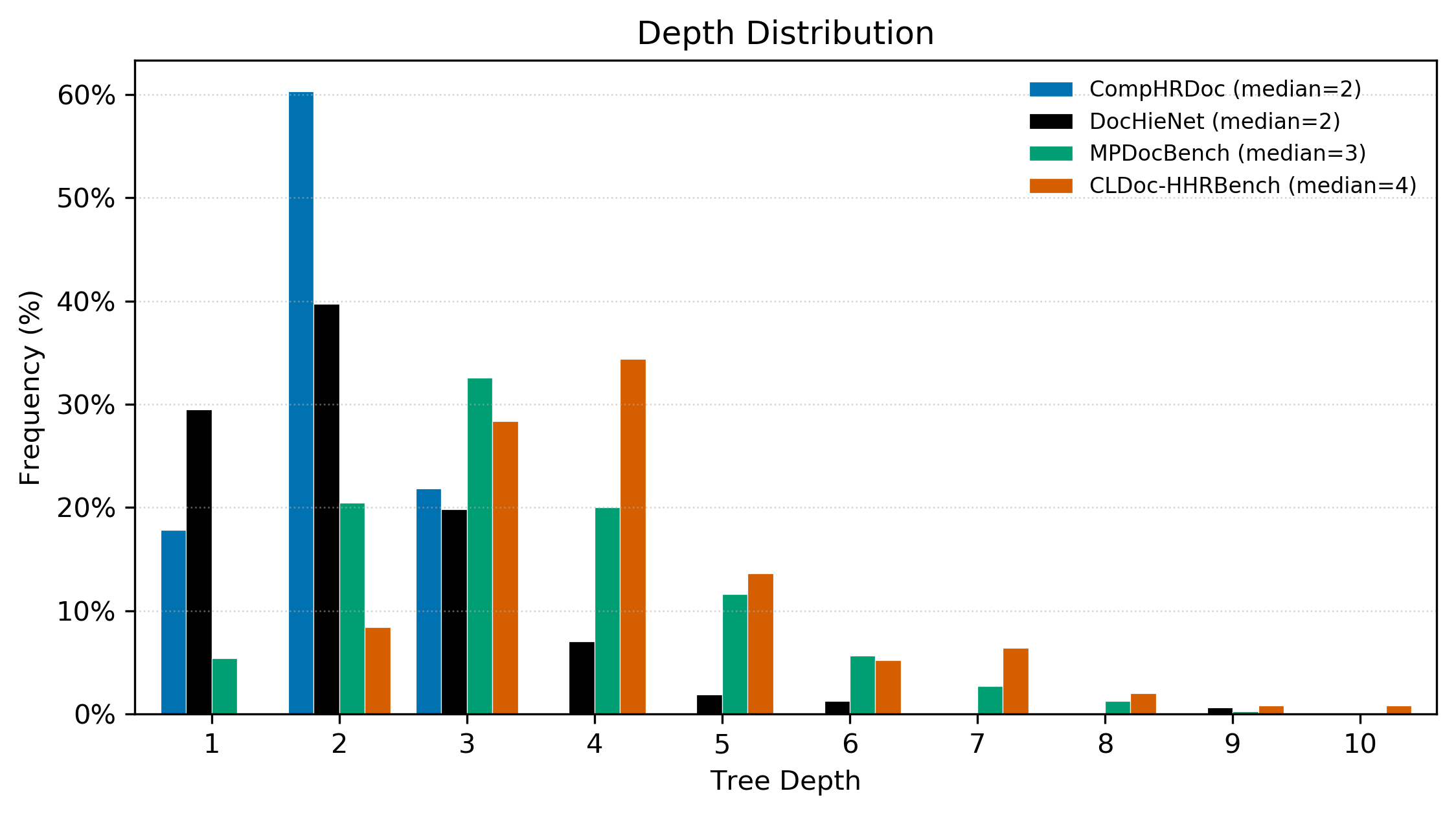}
    \captionof{figure}{Distribution of tree depths across benchmarks. CLDoc-HHRBench has a higher median depth and a broader tail toward deeper hierarchies.}
    \label{fig:depth-distribution}
  \end{minipage}
\end{figure*}

\begin{table*}[!t]
  \centering
  \footnotesize
  \setlength{\tabcolsep}{4pt}
  \renewcommand{\arraystretch}{0.95}
  \begin{tabularx}{\textwidth}{>{\raggedright\arraybackslash}X*{4}{>{\centering\arraybackslash}X}}
    \toprule
    \textbf{Measure} & \textbf{Comp-HRDoc} & \textbf{DocHieNet} & \textbf{MPDocBench-Parse} & \textbf{CLDoc-HHRBench} \\
    \midrule
    \#Docs & 500 & 161 & 420 & 250 \\
    \#Pages (Total) & 7,043 & 2,311 & 3,135 & 11,212 \\
    \#Pages (Mean) & 14.09 & 14.35 & 7.46 & 44.85 \\
    \#Pages (Maximum) & 35 & 50 & 30 & 387 \\
    \#Headings (Total) & 7,579 & 3,344 & 6,309 & 26,231 \\
    \#Headings (Mean) & 15.16 & 20.77 & 15.02 & 104.92 \\
    \#Headings (Maximum) & 64 & 262 & 112 & 1,772 \\
    Depth (Mean) & 2.04 & 2.13 & 3.35 & 4.15 \\
    Depth (Maximum) & 3 & 9 & 9 & 10 \\
    \bottomrule
  \end{tabularx}
  \caption{Comparison with existing datasets. Counts for Comp-HRDoc and DocHieNet refer to their test sets.}
  \label{tab:comparison}
\end{table*}

\subsection{Level Prediction}
We employ a VLM to jointly interpret the visual, textual, and structural information of the headings. Given the memory $M_{i-1}$ and the next heading $h_i$, the VLM predicts its level as
\begin{equation}
    \hat{l}_i = \arg\max_{l\in\mathcal{L}_{i}} p_{\theta}(l \mid M_{i-1}, h_i),
\end{equation}
where $\theta$ denotes the model parameters. For single-heading prediction, the hierarchy level may remain unchanged, increase by one, or return to any shallower level. Accordingly, the candidate set is
\begin{equation}
    \mathcal{L}_{i}=\{1,\ldots,\hat{l}_{i-1}+1\}.
\end{equation}
After predicting $\hat{l}_i=k$, the new heading is attached to the most recent preceding heading at level $k-1$ when $k>1$, or to the virtual root when $k=1$.

Our framework also supports flexible prediction granularity. Because the historical page memory remains fixed within each inference step, the VLM can predict all headings on one or more consecutive target pages using the same memory context. The target pages are provided together with the memory pages, and the VLM outputs the heading levels in reading order. For a group beginning with $h_i$, the first predicted level is constrained by $\hat{l}_{i-1}$, and every subsequent prediction must satisfy $\hat{l}_{j} \leq \hat{l}_{j-1} + 1$. Grouping target pages reduces inference calls and may also provide useful look-ahead context, as later headings in the group can offer visual, semantic, and structural cues for predicting earlier headings.

\begin{table*}[!t]
    \centering
    \scriptsize
    \setlength{\tabcolsep}{2pt}
    \begin{tabularx}{\textwidth}{l|*{2}{>{\centering\arraybackslash}X}|*{2}{>{\centering\arraybackslash}X}|*{2}{>{\centering\arraybackslash}X}|*{2}{>{\centering\arraybackslash}X}|*{2}{>{\centering\arraybackslash}X}}
        \toprule
        \textbf{Methods} & \multicolumn{2}{c|}{\textbf{Comp-HRDoc}} & \multicolumn{2}{c|}{\textbf{DocHieNet}} & \multicolumn{2}{c|}{\textbf{MPDocBench-Parse}} & \multicolumn{2}{c|}{\textbf{CLDoc-HHRBench}} & \multicolumn{2}{c}{\textbf{Overall}} \\
        Name & Macro & Micro & Macro & Micro & Macro & Micro & Macro & Micro & Macro & Micro \\
        \midrule
        \rowcolor[gray]{0.8} \multicolumn{11}{c}{Different Strategies} \\
        \midrule
        PaddleX & \mbox{91.37$\pm$0.00} & \mbox{90.20$\pm$0.00} & \mbox{48.66$\pm$0.00} & \mbox{43.51$\pm$0.00} & \mbox{42.77$\pm$0.00} & \mbox{43.30$\pm$0.00} & \mbox{48.55$\pm$0.00} & \mbox{39.68$\pm$0.00} & \mbox{62.83$\pm$0.00} & \mbox{49.60$\pm$0.00} \\
        MinerU & \mbox{94.33$\pm$0.13} & \mbox{\underline{94.10$\pm$0.03}} & \mbox{74.86$\pm$0.57} & \mbox{73.59$\pm$0.15} & \mbox{71.26$\pm$0.55} & \mbox{71.45$\pm$0.42} & \mbox{80.62$\pm$0.20} & \mbox{73.10$\pm$2.47} & \mbox{82.12$\pm$0.22} & \mbox{76.66$\pm$1.38} \\
        Dynamic Chunking & \mbox{93.08$\pm$0.25} & \mbox{92.38$\pm$0.25} & \mbox{\underline{78.07$\pm$0.19}} & \mbox{75.65$\pm$0.37} & \mbox{70.38$\pm$0.33} & \mbox{70.94$\pm$0.93} & \mbox{80.96$\pm$0.56} & \mbox{\underline{83.61$\pm$0.31}} & \mbox{82.15$\pm$0.09} & \mbox{\underline{82.71$\pm$0.54}} \\
        Pseudo-ToC & \mbox{\underline{94.44$\pm$0.20}} & \mbox{94.04$\pm$0.21} & \mbox{74.61$\pm$0.34} & \mbox{\underline{76.17$\pm$0.20}} & \mbox{\underline{72.94$\pm$0.11}} & \mbox{\underline{75.13$\pm$0.12}} & \mbox{\underline{83.90$\pm$0.33}} & \mbox{79.09$\pm$1.71} & \mbox{\underline{83.64$\pm$0.10}} & \mbox{81.03$\pm$0.98} \\
        Ours & \mbox{\textbf{95.73$\pm$0.26}} & \mbox{\textbf{95.49$\pm$0.19}} & \mbox{\textbf{78.98$\pm$0.41}} & \mbox{\textbf{79.03$\pm$0.40}} & \mbox{\textbf{73.07$\pm$0.62}} & \mbox{\textbf{76.60$\pm$0.32}} & \mbox{\textbf{86.01$\pm$0.24}} & \mbox{\textbf{91.07$\pm$0.15}} & \mbox{\textbf{84.73$\pm$0.32}} & \mbox{\textbf{88.68$\pm$0.17}} \\
        \midrule
        \rowcolor[gray]{0.8} \multicolumn{11}{c}{Separate SFT Results} \\
        \midrule
        MinerU-Popo\textsuperscript{*} (4B) & \mbox{96.32$\pm$0.12} & \mbox{95.86$\pm$0.20} & \mbox{76.66$\pm$0.27} & \mbox{70.97$\pm$0.10} & \mbox{71.87$\pm$0.09} & \mbox{71.13$\pm$0.10} & \mbox{84.18$\pm$0.06} & \mbox{85.64$\pm$0.42} & \mbox{83.95$\pm$0.02} & \mbox{84.06$\pm$0.25} \\
        Ours (0.8B) & \mbox{96.09$\pm$0.25} & \mbox{95.53$\pm$0.25} & \mbox{75.95$\pm$0.54} & \mbox{74.62$\pm$0.56} & \mbox{67.41$\pm$0.15} & \mbox{69.15$\pm$0.13} & \mbox{87.08$\pm$0.27} & \mbox{87.85$\pm$0.46} & \mbox{82.91$\pm$0.12} & \mbox{85.32$\pm$0.25} \\
        \bottomrule
    \end{tabularx}
    \caption{Performance comparison of different methods using Macro-STEDS and Micro-STEDS. Scores are percentages reported as mean $\pm$ standard deviation over three runs, except for deterministic PaddleX, which is evaluated once. Bold and underline indicate the best and second-best results, respectively, in each column within the Different Strategies group. The 0.8B SFT result illustrates the feasibility of fine-tuning our framework on a small model; the separately trained MinerU-Popo\textsuperscript{*} 4B model is shown only as an external reference, not a controlled comparator.}
    \label{tab:main-results}
\end{table*}

\section{Benchmark}
\subsection{Benchmark Construction}
To evaluate heading hierarchy reconstruction on complex and long documents, we construct \textbf{CLDoc-HHRBench}, comprising 250 documents and 11,212 pages. The source files are publicly available PDFs or Microsoft Word documents collected from the Internet, with no document-level overlap with the existing benchmarks used in this work. The benchmark spans 10 document types (Appendix~\ref{sec:benchmark-domains}) and includes long documents and deep heading hierarchies. Figures~\ref{fig:page-distribution} and~\ref{fig:depth-distribution} compare its document-length and tree-depth distributions with existing benchmarks, and Table~\ref{tab:comparison} summarizes their statistics.

\subsection{Annotation}
We identify the headings in each document and annotate their bounding boxes, text, reading order, and hierarchy levels. Each document was labeled independently by two annotators and adjudicated by experts. Six annotators and three experts spent approximately 800 person-hours; the annotation protocol and validation are described in Appendix~\ref{sec:annotation-protocol}.

\section{Experiments}
\subsection{Experiment Setup} 
We compare our method with four baselines. The first is \textbf{PaddleX}\footnote{\url{https://github.com/PaddlePaddle/PaddleX/blob/release/3.7/paddlex/inference/pipelines/layout_parsing/title_level.py}}, which constructs the heading hierarchy using rule-based features, including heading text, numbering patterns, bounding-box height, and global reading order. The second is \textbf{MinerU}\footnote{\url{https://github.com/opendatalab/MinerU/blob/master/mineru/utils/llm_aided.py}}, which employs an LLM to infer heading levels by jointly processing the ordered headings across the entire document. The remaining two are \textbf{Dynamic Chunking (MinerU-Popo)} \cite{xu2026mineru} and \textbf{Pseudo-ToC (LingDT-VL-OCR)} \cite{qian2026lingdtvlocr}.

A uniform comparison using the original implementations of both VLM-based baselines is not feasible: MinerU-Popo fine-tunes Qwen3-VL-4B\footnote{\url{https://huggingface.co/Qwen/Qwen3-VL-4B-Instruct}} with 30K training instances, whereas LingDT-VL-OCR has not released its model. We therefore reproduce their heading hierarchy reconstruction strategies with Qwen3.6-Plus\footnote{\url{https://qwen.ai/blog?id=qwen3.6}} as a shared backbone, thereby isolating differences among the reconstruction strategies. Apart from strategy-specific prompts, all reproduced VLM-based strategies use the same backbone and inference hyperparameters. Unless otherwise specified, our method uses one-page prediction. An output is accepted only if it can be parsed as valid JSON and satisfies $1 \leq \hat{l}_i \leq \hat{l}_{i-1}+1$ for every target in reading order. If validation fails, we retry the prediction up to two times. For the rare case in which a single-heading prediction remains invalid, we project the predicted level onto the valid range.

To assess whether our framework can be used to fine-tune a small model, we conduct a separate SFT feasibility experiment. We collect documents from the training sets of Comp-HRDoc and DocHieNet and from publicly available documents found online. We manually annotate them to create 10K training instances and fine-tune Qwen3.5-0.8B\footnote{\url{https://huggingface.co/Qwen/Qwen3.5-0.8B}} using our method. The documents underlying these training instances are strictly disjoint from all evaluation documents. We train with Llama-Factory \cite{zheng2024llamafactory} on three NVIDIA A100 GPUs for one epoch, using a learning rate of $5.0 \times 10^{-6}$, a per-device batch size of 2, four gradient-accumulation steps, and an 8K maximum context length.

For each reproduced baseline, we follow the default settings specified in its paper or official GitHub repository. The experimental prompts and remaining implementation details are provided in Appendix~\ref{sec:appendix-prompts} and Appendix~\ref{sec:appendix-implementation}.

\begin{figure}[!t]
    \centering
    \includegraphics[width=\linewidth]{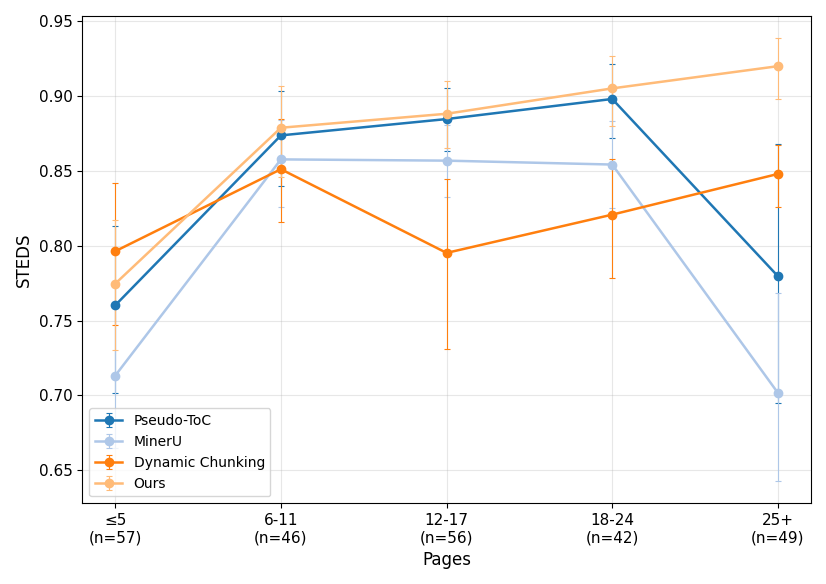}
    \caption{Micro-STEDS by the number of heading-bearing pages on CLDoc-HHRBench. Error bars show 95\% document-bootstrap confidence intervals; $n$ denotes documents per group.}
    \label{fig:steds-pages-all}
    \vspace{6pt}
    \includegraphics[width=\linewidth]{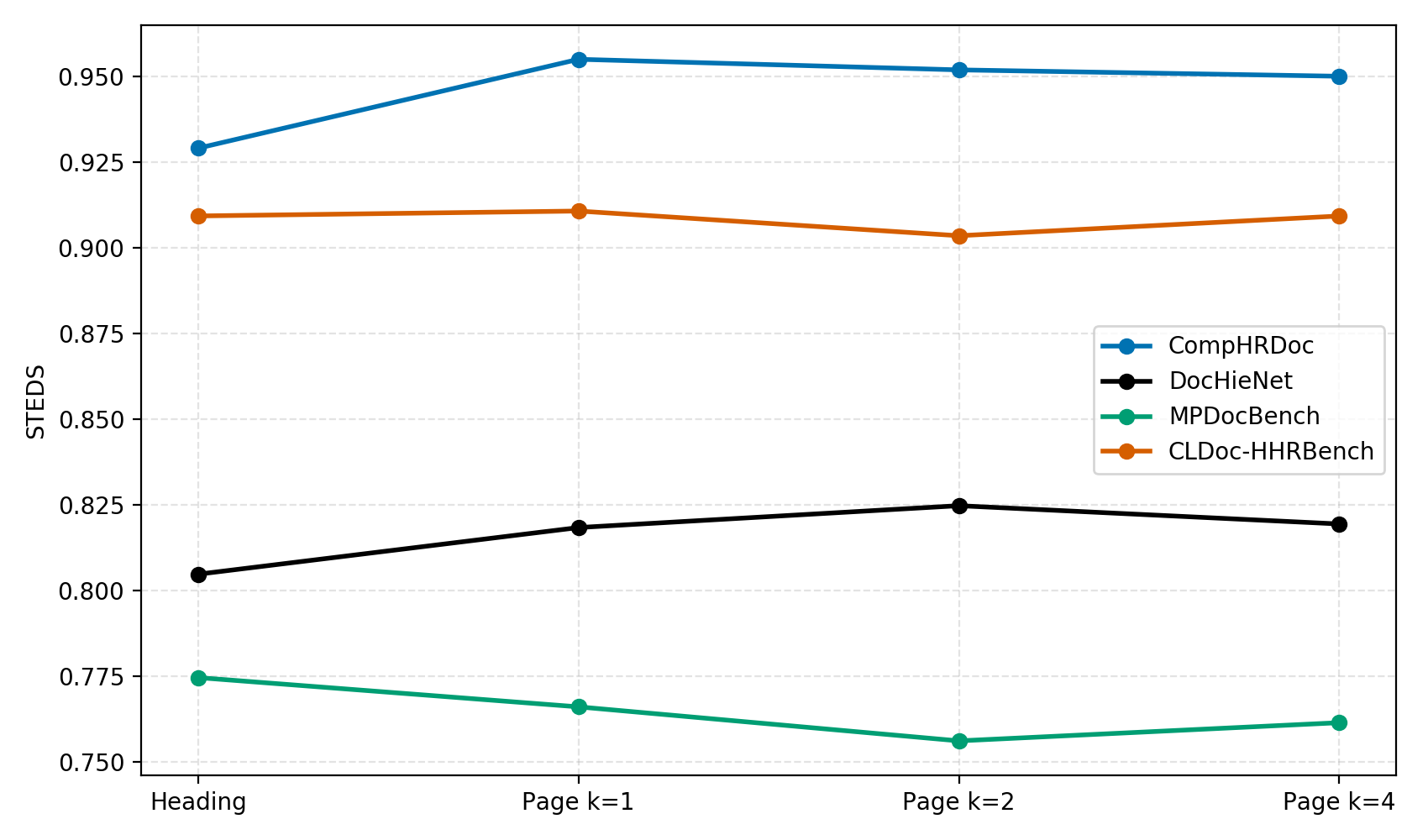}
    \caption{Effect of prediction granularity on Micro-STEDS across four benchmarks.}
    \label{fig:prediction-granularity}
\end{figure}

\subsection{Evaluation Protocols} 
We compare hierarchy reconstruction strategies using the same ground-truth headings for every benchmark document. All methods receive annotated heading text, page indices, bounding boxes, and reading order, and predict only hierarchy levels. In the main comparison, ground-truth hierarchy levels are not provided as historical context: every incremental method updates its state using its own previous level predictions. This evaluation measures the hierarchy reconstruction strategies themselves; the impact of upstream errors on an end-to-end document parsing pipeline is outside the scope of this work.

Following prior work \cite{ma2023hrdoc}, we adopt STEDS as the primary evaluation metric. For document $i$ among the $N$ evaluated documents, let $T_i$ and $\hat{T}_i$ denote its ground-truth and predicted hierarchies, respectively, where $ed_i=\mathrm{EditDist}(T_i,\hat{T}_i)$ is their tree edit distance and $n_i=\max(|T_i|,|\hat{T}_i|)$ is the larger number of nodes in the two trees. We report Macro-STEDS and Micro-STEDS:
\begin{equation}
    \mathrm{Macro\mbox{-}STEDS}=\frac{1}{N}\sum_{i=1}^{N}\left(1-\frac{ed_i}{n_i}\right),
\end{equation}
\begin{equation}
    \mathrm{Micro\mbox{-}STEDS}=1-\frac{\sum_{i=1}^{N}ed_i}{\sum_{i=1}^{N}n_i}.
\end{equation}
A Micro-STEDS score higher than the corresponding Macro-STEDS score generally indicates better performance on longer documents or larger hierarchy trees. Unless otherwise specified, we report Micro-STEDS.

\subsection{Experimental Results}

Table~\ref{tab:main-results} demonstrates the overall effectiveness of our method. In the comparison of different strategies, it achieves Overall Macro-STEDS and Micro-STEDS scores of 84.73 and 88.68, outperforming the strongest competing results by 1.09 and 5.97 points, respectively. The advantage is most pronounced on CLDoc-HHRBench, where our score of 91.07 exceeds the next-best score of 83.61 by 7.46 points, demonstrating the effectiveness of our method on long and structurally complex documents. In contrast, PaddleX remains competitive only on Comp-HRDoc and degrades markedly on the other benchmarks, indicating the limited generalization of rule-based methods. MinerU, which relies on textual heading sequences, underperforms the multimodal VLM-based methods overall, suggesting that page images provide more comprehensive visual and structural cues. Dynamic Chunking also performs worse overall, likely because vertical page concatenation weakens visual cues and overlap-based synchronization cannot fully correct non-uniform errors.

In the separate SFT feasibility experiment, fine-tuning Qwen3.5-0.8B with our framework yields 85.32 Overall Micro-STEDS, showing that the framework can also be applied to a small model.

Figure~\ref{fig:steds-pages-all} groups CLDoc-HHRBench documents by the number of heading-bearing pages and reports their Micro-STEDS scores. Our advantage is most pronounced for documents with at least 25 such pages. As documents lengthen, Pseudo-ToC must process more page images and generate a longer hierarchy, making it harder for the VLM to integrate visual evidence and maintain consistent heading levels. Our level-wise memory selects historical pages by observed hierarchy level and keeps the context compact as page count grows. This may be one reason for our method's larger gains on CLDoc-HHRBench.

\begin{table}[!t]
    \centering
    \scriptsize
    \setlength{\tabcolsep}{0pt}
    \begin{tabularx}{\linewidth}{l*{5}{>{\centering\arraybackslash}X}}
        \toprule
        \textbf{Strategy} & \shortstack{\textbf{Comp-}\\\textbf{HRDoc}} & \textbf{DocHieNet} & \shortstack{\scalebox{0.96}[1]{\textbf{MPDocBench-}}\\\textbf{Parse}} & \shortstack{\textbf{CLDoc-}\\\textbf{HHRBench}} & \textbf{Overall} \\
        \midrule
        Random & \mbox{94.62$\pm$0.14} & \mbox{74.86$\pm$0.24} & \mbox{73.53$\pm$0.61} & \mbox{74.28$\pm$1.77} & \mbox{78.22$\pm$0.75} \\
        Previous-$n$ & \mbox{95.08$\pm$0.07} & \mbox{78.40$\pm$0.20} & \mbox{75.90$\pm$0.34} & \mbox{89.40$\pm$0.21} & \mbox{87.47$\pm$0.12} \\
        Recent & \mbox{\textbf{95.49$\pm$0.19}} & \mbox{\textbf{79.03$\pm$0.40}} & \mbox{\textbf{76.60$\pm$0.32}} & \mbox{\textbf{91.07$\pm$0.15}} & \mbox{\textbf{88.68$\pm$0.17}} \\
        \bottomrule
    \end{tabularx}
    \caption{Micro-STEDS comparison of historical memory selection strategies. Previous-$n$ uses $n=3$.}
    \label{tab:memory-strategies}

    \vspace{8pt}
    \setlength{\tabcolsep}{0pt}
    \begin{tabularx}{\linewidth}{l*{5}{>{\centering\arraybackslash}X}}
        \toprule
        \textbf{Method} & \shortstack{\textbf{Comp-}\\\textbf{HRDoc}} & \textbf{DocHieNet} & \shortstack{\scalebox{0.96}[1]{\textbf{MPDocBench-}}\\\textbf{Parse}} & \shortstack{\textbf{CLDoc-}\\\textbf{HHRBench}} & \textbf{Overall} \\
        \midrule
        Text-only & \mbox{91.23$\pm$0.62} & \mbox{70.96$\pm$0.61} & \mbox{66.78$\pm$0.84} & \mbox{86.71$\pm$0.32} & \mbox{83.47$\pm$0.12} \\
        Ours & \mbox{\textbf{95.49$\pm$0.19}} & \mbox{\textbf{79.03$\pm$0.40}} & \mbox{\textbf{76.60$\pm$0.32}} & \mbox{\textbf{91.07$\pm$0.15}} & \mbox{\textbf{88.68$\pm$0.17}} \\
        \bottomrule
    \end{tabularx}
    \captionof{table}{Micro-STEDS comparison between the text-only variant and our multimodal method.}
    \label{tab:text-only}

    \vspace{8pt}
    \setlength{\tabcolsep}{0pt}
    \begin{tabularx}{\linewidth}{l*{5}{>{\centering\arraybackslash}X}}
        \toprule
        \textbf{Memory} & \shortstack{\textbf{Comp-}\\\textbf{HRDoc}} & \textbf{DocHieNet} & \shortstack{\scalebox{0.96}[1]{\textbf{MPDocBench-}}\\\textbf{Parse}} & \shortstack{\textbf{CLDoc-}\\\textbf{HHRBench}} & \textbf{Overall} \\
        \midrule
        Predicted & \mbox{95.51$\pm$0.17} & \mbox{71.82$\pm$1.07} & \mbox{73.24$\pm$0.18} & \mbox{84.84$\pm$0.87} & \mbox{83.89$\pm$0.49} \\
        GT-guided & \mbox{\textbf{97.25$\pm$0.08}} & \mbox{\textbf{77.41$\pm$0.35}} & \mbox{\textbf{73.38$\pm$0.04}} & \mbox{\textbf{91.59$\pm$0.15}} & \mbox{\textbf{88.73$\pm$0.11}} \\
        \bottomrule
    \end{tabularx}
    \captionof{table}{Parent accuracy (\%) with predicted or fully ground-truth-guided memory.}
    \label{tab:gt-anchors}
\end{table}

\subsection{Analysis}

\subsubsection{Effect of Prediction Granularity}
As our framework supports prediction at multiple granularities, we evaluate four settings under the same evaluation protocol. The prediction unit is set to a single heading, all headings on one page, or all headings on two or four consecutive pages. Figure~\ref{fig:prediction-granularity} reports the results on four benchmarks. Performance remains relatively stable when a prediction unit spans multiple consecutive pages. These results suggest that the framework can accommodate multi-page prediction, which may benefit from look-ahead context, although its effect and the optimal granularity vary across benchmarks.

\subsubsection{Comparison of Memory Strategies}
Table~\ref{tab:memory-strategies} compares our method with Previous-$n$ and Random on four benchmarks. Previous-$n$ retains the $n$ headings immediately preceding the current target in reading order as historical context, whereas our Most Recent strategy selects the latest heading at each observed level. Random selects one processed node at each observed level at random. Overall, Most Recent achieves a Micro-STEDS of 88.68, compared with 87.47 for Previous-$n$ and 78.22 for Random, underscoring the importance of selecting useful historical context.

\subsubsection{Effect of Multimodal Context}
To assess the contribution of visual information, we compare our method with a text-only variant under the same evaluation protocol. The text-only variant follows the same memory selection and prediction procedure but excludes page images, while retaining heading texts, bounding-box coordinates, and predicted levels in textual form. Table~\ref{tab:text-only} reports the results on four benchmarks. Our multimodal method consistently outperforms the text-only variant across all benchmarks, demonstrating the importance of visual context for reconstructing document hierarchies.

\subsubsection{Sensitivity to Historical Hierarchy Information}
Because both anchor selection and the level labels exposed in memory depend on the predicted hierarchy, historical errors may affect subsequent predictions. We examine this sensitivity with a fully ground-truth-guided diagnostic: the ground-truth hierarchy is used to select the most recent anchor at each level, and all historical level labels exposed in the resulting memory are replaced with their ground-truth values; all other settings remain unchanged. Table~\ref{tab:gt-anchors} reports parent accuracy, the proportion of headings attached to the correct parent. The predicted-memory setting already achieves 83.89\% overall parent accuracy, while replacing the entire historical state with ground-truth information improves it to 88.73\%. This moderate overall gap suggests that imperfect historical predictions do not cause severe cascading degradation at the aggregate level.

\section{Conclusion}
We introduce level-wise multimodal memory for VLM-based heading hierarchy reconstruction. Retaining the most recent heading at each observed level provides compact visual and structural context for heading-level and multi-page prediction. We also present CLDoc-HHRBench, a diverse benchmark of long documents and deep hierarchies. Experiments on four benchmarks show particularly strong results on complex, long documents.

\section*{Ethical Considerations}
CLDoc-HHRBench is sourced from publicly available PDF and Microsoft Word documents found online and manually annotated by our annotators, with expert adjudication. We have secured the necessary permissions to include the source documents in the benchmark. CLDoc-HHRBench carries low ethical risk: we screen the documents and annotations to exclude sensitive or personally identifiable information. We will release the annotations and source information for the page images.

\bibliography{references}

\clearpage
\appendix

\makeatletter
\setlength{\@fptop}{0pt}
\setlength{\@fpbot}{0pt plus 1fil}
\makeatother

\newtcblisting{classicprompt}[1]{
  listing only,
  colback=gray!4,
  colframe=gray!48,
  boxrule=0.55pt,
  arc=2pt,
  left=3pt,
  right=3pt,
  top=1pt,
  bottom=1pt,
  title={#1},
  fonttitle=\bfseries\sffamily\fontsize{8pt}{8pt}\selectfont,
  coltitle=black,
  listing options={
    basicstyle=\fontsize{7pt}{8pt}\selectfont\ttfamily,
    numbers=none,
    xleftmargin=0pt,
    breaklines=true,
    columns=fullflexible,
    keepspaces=true,
    showstringspaces=false,
    literate={λ}{{$\lambda$}}1 {ξ}{{$\xi$}}1
  }
}

\section{Benchmark Domain Breakdown}
\label{sec:benchmark-domains}
The benchmark covers 10 document types: Insurance Policy Terms (10.0\%), Product Manuals (10.8\%), Laws and Regulations (5.2\%), Financial Reports (12.0\%), Guidelines (3.6\%), Prospectuses (3.6\%), Rating Reports (3.6\%), Academic Papers (4.0\%), Industry Standards (3.6\%), and Research Reports (43.6\%). Research Reports cover documents that differ markedly in layout, length, and heading structure. This category can be further divided into five subcategories: In-Depth Research (11.2\% of all documents), Initiation of Coverage (8.8\%), Company Update (13.2\%), Sector Research (5.6\%), and \mbox{Macro \& Strategy Research} (4.8\%). 
\section{Annotation Protocol}
\label{sec:annotation-protocol}
CLDoc-HHRBench was manually annotated from the original pages using a shared protocol. We first identify which elements in each document are headings, then annotate each heading's bounding box, text, reading order, and normalized hierarchy level. Skipped levels were prohibited: a heading could be at most one level deeper than its predecessor. We also developed a visual annotation platform that displayed pages alongside the hierarchy tree and supported direct editing of all targets.

Each document was independently labeled by two mutually blinded annotators. Before adjudication, agreement was 0.9611 F1 for heading detection, 0.9334 mean box IoU, 0.9761 normalized edit similarity for reading order, and 0.9673 hierarchy STEDS \cite{ma2023hrdoc}. The two complete annotation sets for each document were then compared. One of the three experts reviewed each missing or extra heading and each mismatch in box, text, order, or level against the original pages. Any remaining ambiguities were resolved by all three experts through consensus.

After adjudication, deterministic validators checked reading order indices, out-of-page boxes, empty text, skipped levels, and sequence--tree consistency. Flagged documents were corrected and revalidated until all checks passed. Six annotators and three experts spent approximately 800 person-hours. We will release the annotations and source information for the page images.

\section{Prompts}
\label{sec:appendix-prompts}
This section provides the prompts used in our experiments, including the prompt of our method and the strategy-specific prompts of the reproduced VLM-based baselines. The prompts of our method, Pseudo-ToC (LingDT-VL-OCR), and Dynamic Chunking (MinerU-Popo) are shown below, each instantiated with headings from the same example document.

In each prompt, \texttt{<image>} denotes the visual input of the corresponding strategy: our method provides the original page images with the heading bounding boxes drawn on them, LingDT-VL-OCR provides the pseudo-ToC images aggregated from heading crops, and MinerU-Popo provides the single image formed by vertically concatenating the pages within a chunk. All strategies share the same output format: the VLM outputs JSON only, mapping each target heading id to its hierarchy level. Because MinerU-Popo releases a fine-tuned model that follows its own output format, the original Dynamic Chunking prompt does not include the output instruction; we append it only when reproducing the strategy with the shared backbone, leaving the rest of the prompt unchanged.

For different prediction granularities, the prompt differs only in the number of target headings to be predicted and the pages containing them. For the alternative memory-selection strategies, Previous-$n$ and Random, the prompt is similar to that of our method: only the given page images and the corresponding heading information are replaced according to the selected historical nodes, and everything else remains unchanged. For the text-only variant, the prompt differs only in that the page images are excluded and the bounding boxes are serialized in textual form. The ground-truth-guided diagnostic also reuses the prompt of our method: the given page images and the historical levels exposed in the memory are replaced according to the ground-truth hierarchy. The SFT model is fine-tuned and evaluated with the same prompt used by our method.

\begin{figure}[!t]
\begin{classicprompt}{Prompt used for Our Method}
Parse the titles' hierarchy.

Page Images (each image is preceded by its id [i]):

[1] <image>

[2] <image>

Known Title Levels:
[{"title": "I. INTRODUCTION", "level": "H1", "image_id": 1}, {"title": "II. RGE FOR λH BELOW SCALE MP /ξ", "level": "H1", "image_id": 1}]

Titles to Predict Hierarchy Level:
[{"id": 0, "title": "III. EMBEDDING HIGGS INFLATION INTO", "image_id": 2}, {"id": 1, "title": "SUPERGRAVITY", "image_id": 2}]

Output JSON only, mapping each id to its level, e.g. {"0": "H1", "1": "H2"}. No explanation.
\end{classicprompt}

\begin{classicprompt}{Prompt used for Pseudo-ToC (LingDT-VL-OCR)}
Parse the titles' hierarchy.

Page Images (each image is preceded by its id [i]):

[1] <image>

Titles to Predict Hierarchy Level:

[{"id": 0, "title": "I. INTRODUCTION"}, {"id": 1, "title": "II. RGE FOR λH BELOW SCALE MP /ξ"}, {"id": 2, "title": "III. EMBEDDING HIGGS INFLATION INTO"}, {"id": 3, "title": "SUPERGRAVITY"}, {"id": 4, "title": "IV. CONSTRAINTS ON SUSY MASS SPECTRUM"}, {"id": 5, "title": "V. DISCUSSION AND SUMMARY"}, {"id": 6, "title": "ACKNOWLEDGMENTS"}]

Output JSON only, mapping each id to its level, e.g. {"0": "H1", "1": "H2"}. No explanation.
\end{classicprompt}

\begin{classicprompt}{Prompt used for Dynamic Chunking (MinerU-Popo)}
<image>

Title Level Analysis: <|id|>0<|page|>0<|box|>239 220 250 351<|content|>I. INTRODUCTION
<|id|>1<|page|>0<|box|>825 586 837 848<|content|>II. RGE FOR λH BELOW SCALE MP /ξ
<|id|>2<|page|>1<|box|>406 570 417 864<|content|>III. EMBEDDING HIGGS INFLATION INTO
<|id|>3<|page|>1<|box|>420 660 431 774<|content|>SUPERGRAVITY
<|id|>4<|page|>2<|box|>68 555 79 879<|content|>IV. CONSTRAINTS ON SUSY MASS SPECTRUM
<|id|>5<|page|>3<|box|>808 176 819 398<|content|>V. DISCUSSION AND SUMMARY
<|id|>6<|page|>3<|box|>476 637 487 797<|content|>ACKNOWLEDGMENTS

Output JSON only, mapping each id to its level, e.g. {"0": "H1", "1": "H2"}. No explanation.
\end{classicprompt}
\end{figure}

\section{Implementation Details}
\label{sec:appendix-implementation}
This section describes the remaining implementation details, including method-specific configurations and retry and validation settings. The method-specific configurations are summarized as follows:
\begin{itemize}
    \item \textbf{PaddleX}: We reproduce it using the source code at \url{https://github.com/PaddlePaddle/PaddleX/blob/release/3.7/paddlex/inference/pipelines/layout_parsing/title_level.py}.
    \item \textbf{MinerU}: We reproduce it using the source code at \url{https://github.com/opendatalab/MinerU/blob/master/mineru/utils/llm_aided.py}, with Qwen3.6-Plus as the LLM.
    \item \textbf{Dynamic Chunking}: We follow \url{https://github.com/opendatalab/MinerU-Popo} with its default settings, including a chunk size of 50 and an overlap of 1.
    \item \textbf{Pseudo-ToC}: For the construction of the pseudo-ToC images, each canvas is white and has the same size as the first page containing a heading. Each heading crop is preceded by its rendered id and keeps its original horizontal position; the crops are tiled densely along the vertical axis from top to bottom with a fixed 12px gap, and a new blank canvas is started when the current one overflows.
    \item \textbf{Our Method}: The heading bounding boxes are drawn directly on the original page images, and all page images are provided at their original size.
\end{itemize}

We apply the same output validation and retry procedure to MinerU, Dynamic Chunking, Pseudo-ToC, and our method. Each output must be valid JSON, and every predicted heading level must satisfy $1 \leq \hat{l}_i \leq \hat{l}_{i-1}+1$, since the four benchmarks contain no skipped levels. We retry invalid predictions up to two times. If a parseable single-heading prediction still violates the level constraint, we project it onto the valid range: $\hat{l}_i \leftarrow \min(\max(\hat{l}_i,1),\hat{l}_{i-1}+1)$.

\end{document}